# An explainable Transformer-based deep learning model for the prediction of incident heart failure

Shishir Rao, MSc[1,2,†], Yikuan Li, MSc[1,2,†], Rema Ramakrishnan, PhD[1,2], Abdelaali Hassaine, PhD[1,2], Dexter Canoy, PhD [1,2,3], John Cleland, MD, FRCP, FESC[4], Thomas Lukasiewicz, PhD[5], Gholamreza Salimi-Khorshidi, PhD[1,2], Kazem Rahimi, DM, FESC[1,2,3] *

[1] Deep Medicine, Oxford Martin School, University of Oxford, Oxford, United Kingdom

[2] Nuffield Department of Women's & Reproductive Health, University of Oxford, Oxford, United Kingdom

[3] NIHR Oxford Biomedical Research Centre, Oxford University Hospitals NHS Foundation Trust, Oxford, United Kingdom

[4] Robertson Centre for Biostatistics, University of Glasgow, Glasgow, United Kingdom

[5] Department of Computer Science, University of Oxford, Oxford, United Kingdom

[†]These authors contributed equally to this work

[*]Corresponding author: Kazem Rahimi, Deep Medicine, Oxford Martin School, University of Oxford, Oxford, United Kingdom; email address: kazem.rahimi@wrh.ox.ac.uk; Tel: +44 (0) 1865 617200; Fax: +44 (0) 1865 617202; twitter handle: kazemr

# Abstract

Predicting the incidence of complex chronic conditions such as heart failure is challenging. Deep learning models applied to rich electronic health records may improve prediction but remain unexplainable hampering their wider use in medical practice. We developed a novel Transformer deep-learning model for more accurate and yet explainable prediction of incident heart failure involving 100,071 patients from longitudinal linked electronic health records across the UK. On internal 5-fold cross validation and held-out external validation, our model achieved 0.93 and 0.93 area under the receiver operator curve and 0.69 and 0.70 area under the precision-recall curve, respectively and outperformed existing deep learning models. Predictor groups included all community and hospital diagnoses and medications contextualized within the age and calendar year for each patient's clinical encounter. The importance of contextualized medical information was revealed in a number of sensitivity analyses, and our perturbation method provided a way of identifying factors contributing to risk. Many of the identified risk factors were consistent with existing knowledge from clinical and epidemiological research but several new associations were revealed which had not been considered in expert-driven risk prediction models.

# Abbreviations

EHR: electronic health records

GP: general practices

HF: heart failure

# Introduction

Heart failure (HF) remains a major cause of morbidity, mortality, and economic burden.[1] Despite recent evidence suggesting improvements in the quality of clinical care that patients with HF receive, and favorable trends in prognosis,[2] the incidence of HF has changed little.[3] Indeed, as a consequence of population growth and ageing, the absolute burden of HF has been increasing, with incidence rates similar to the four most common causes of cancer combined.[3] These observations reinforce the need for fuller implementation of existing strategies for HF prevention and further investigations into risk factors. Several statistical models have been developed to predict risk of incident HF; however, the predictive performance of these models has been largely unsatisfactory.[1]

The growing availability of comprehensive clinical datasets, such as linked electronic health records (EHR) with extensive clinical information from a large number of individuals, together with advances in machine learning, offer new opportunities for developing more robust risk-prediction models than conventional statistical approaches.[4,5] Such data-driven approaches can also potentially discover new associations that are less dependent on expert knowledge. However, empirical evidence of robust prediction of complex chronic conditions, such as HF, has been limited. Prominent deep learning architectures have only shown modest performance in large-scale, complex EHR datasets.[6] Furthermore, due to their high level of abstraction, these deep learning models have typically had poor explainability, which limits their trustfulness and contribution to risk factor discoveries and wider application in clinical settings.[7]

In this study, we aimed to develop and validate a model for predicting incident HF, leveraging state-of-the-art, deep-sequential architecture applied to temporal and multi-modal EHR.

## Methods

### Dataset

We used UK Clinical Practice Research Datalink (CPRD), one of the largest de-identified longitudinal population based-EHR databases nationally representative in terms of age, sex, and ethnicity.[8] This was used with approval from International Scientific Advisory Committee (ISAC protocol: 17_224R2). It contains primary care data from general practices (GP) since 1985, and links to secondary care and other health and area-based administrative databases[4,8] (e.g. Hospital Episode Statistics[9]).

We selected GP records that met certain quality standards for research (as assessed by CPRD) and those with full data-linkage with secondary care to retain a complete patient history between 1985 and 2015. Male and female patients were included with records from 16 years of age. Afterwards, a two-step process was applied to select a cohort for general representation pre-training and a sub-cohort for incident HF prediction (Figure 1). Here, incident HF is defined as the first HF diagnosis code for each patient during the study period. Firstly, we included patients with at least five visits to ensure sufficient contextual information for representation learning. This cohort included 1,609,024 patients in total and is referred to as dataset A. Secondly, to develop models for predicting incident HF, we selected a subset of dataset A with richer medical information. More specifically, we kept patients with i) at least 10 visits to their GP or hospital, ii) at least three years of records, iii) at least 10 unique codes been recorded. For patients with at least one diagnosis of HF, information up to six months before the first record of HF was used for learning; our process ensures no HF diagnoses in learning period. For each patient without HF, we randomly select a time stamp and consider all records prior as the learning period. This led to selection of a cohort of 100,071 patients, with 13,050 (13%) cases of incident HF, henceforth referred to as dataset B.

## Model architecture

We used the BEHRT model architecture reported by Li et al[10] inspired by Transformer[11] models and extended it to meet the study objectives. In brief, the model captures disease and medication associations within their temporal context (Figure 1E), to bolster predictive performance. BEHRT works robustly with large-scale, sequential data and outperforms other classical machine and deep learning models on subsequent visit prediction tasks (appendix section 1 for more information).[10]

We included encounter (disease and medication), age, and calendar year as input information. Each of the three modalities are represented by a trainable embedding matrix[10], which is a two-dimensional matrix with each instance as a vector (Figure 1C). Each encounter with its respective age and calendar year layers are summed to form a single predictor in the model.

## Statistical analysis

To assess and validate model performance for HF prediction, we selected 60%, 20%, and 20% of the patients in dataset B as training, testing, and external, held-out validation cohorts, respectively. Five-fold cross-validation was applied for internal validation. Model performance was assessed using area under the receiver operating characteristic (AUROC) and precision-recall curve (AUPRC) with 95% confidence interval (CI) over five folds.

We conducted ablation study to assess importance of different data modalities (diagnoses (D), medications (M), age (A), and calendar year (Y)) by alternatively removing each of them. More specifically, we devised six experiments with the following combination of modalities as input into the models: D, DA, DAY, DM, DMA, and DMAY, with letters corresponding to modalities; also, a visualized analysis for time-related modalities (i.e. age and year) is included in appendix section 2. We also replicated state-of-the-art deep learning model, RETAINEX[12], which has outperformed other models such as RETAIN, logistic regression, Deepr, Deep Patient, and eNRBM[12] on our dataset and compared it directly with our model.

Next, we aimed to develop ways of quantifying encounter contributions to the prediction of incident HF, as a way of making models explainable. For this, we used a perturbation technique inspired by work of Guan et al[13] on the summed, predictor embeddings to represent the disease or medication as seen in Figure 1H. The fundamental concept is to measure change in predictive probability after perturbing the input to indicate the contribution of predictors. If large perturbations of predictors minimally change outcome probability, then the predictors are unimportant for prediction. However, if minimal perturbation greatly changes outcome probability, the respective predictors are highly important. In this work, we proposed an asymmetric loss function to prioritize encounters that enhanced HF/non-HF predictions. More details can be found in appendix section 3.

In case of repeat diagnoses/prescriptions for the same patient, we only considered the contribution for the first recorded instance. We calculate the average contribution for a particular disease/medication across all HF and non-HF patients (Figure 1H). We analyzed predictors where their respective disease codes occurred in at least 5% of patients to ensure sufficient training of the input variables. Additionally, we considered established risk factors from statistical models for HF prediction even if prevalence was less than 5%.[14,15] We calculated the relative contribution (RC) with 95% CI[16] of a particular disease/medication by dividing the average contribution in HF patients by the average contribution in non-HF patients with value greater than 1.0 and less than 1.0 implying that the disease/medication is positively associated with HF and non-HF respectively.

To understand differential contribution to HF by age and calendar year, we repeated the encounter contribution analyses stratified by age groups: (50 to 60], (60 to 65], (65 to 70], (70 to 75], and (75 to 80] and calendar year groups: [1990, 1995], (1995-2000], (2000-2005], (2005-2010] when clinical events were first recorded.

This analysis was conducted on a subset of patients from the external held-out testing and validation datasets where BEHRT (DMAY) performed best named Dataset C (appendix section 4).

# Results

## Dataset preparation

Our dataset included diagnostic codes (299 codes in Caliber[17]) and medications (426 codes) at each encounter as well as patient age in months and calendar year. Both disease and medication codes with unknown mapping were mapped to an "UNKNOWN" category. Of the 100,071 patients for incident HF prediction (dataset B), the median age at baseline was 70 years, 52% were men, 65% had history of hypertension, 9% a prior myocardial infarction, and 5.1% an ischemic stroke. The median duration of follow-up (date of the first record to baseline) was 9 years. More details for training, test, and external validation set, as well as code phenotyping and specifically, heart failure phenotypying processes can be found in Table S1, Figure S1, and appendix section 5.

## Model performance

We implemented BEHRT to predict incident HF[10]. Our model, with inclusion of all four modalities (DMAY), showed the best performance for incident HF prediction, with an AUROC of 0.93 and AUPRC of 0.70 in the external validation cohort (Figure 2). A comparison of our model with RETAINEX[12] (in the full cohort) showed noticeable improvement in predictive ability – 3%/9% absolute improvement in AUROC/AUPRC (Figure 2). Figure 2 also shows that model with calendar year as the contextualizing variable demonstrates substantial improvements in terms of AUROC/AUPRC than one with age, thus, indicating calendar year to be more informative for contextualizing predictors than chronological age (visualized analysis found in appendix section 6). Lastly, after evaluating BEHRT (DMAY) on combined test and validation of external validation dataset, we found the best performing quintiles of predictive probability were the first and fifth with quintile-AUROC of 0.84 and 0.71, respectively.

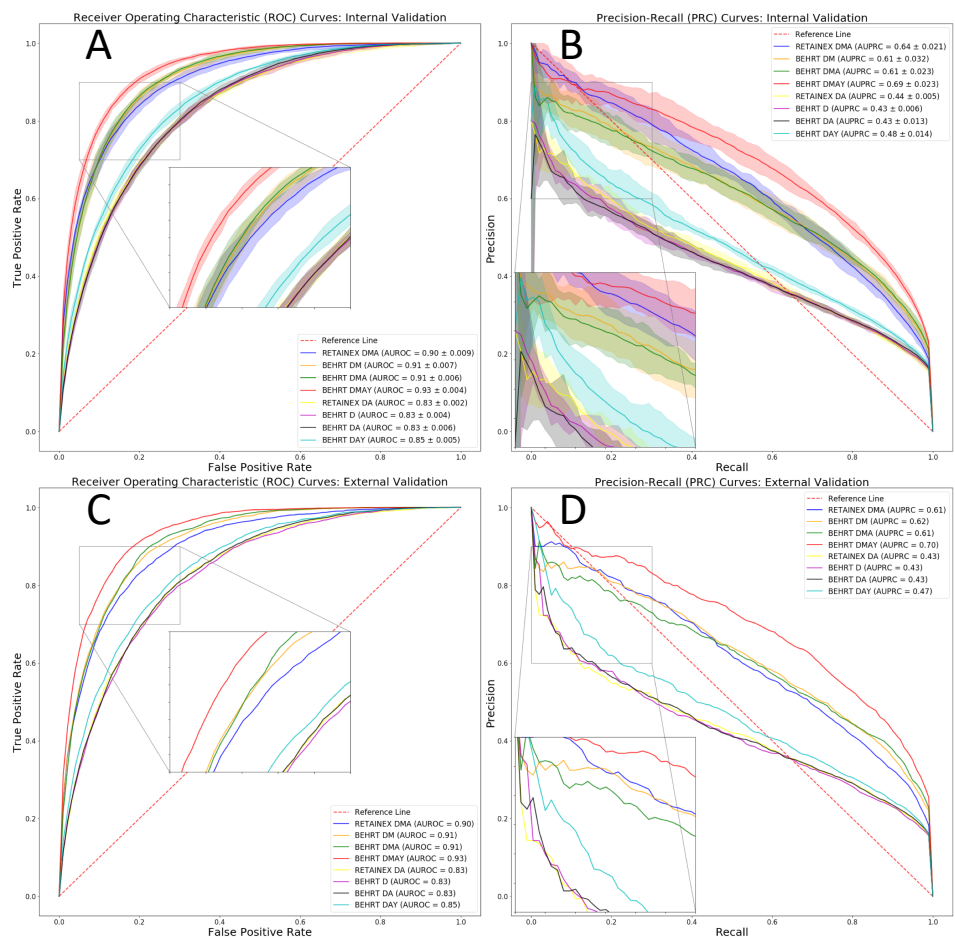

**Figure 2. Model performance evaluation and temporal modality analysis**. A and B represent 5-fold internal validation; C and D represent external validation. A and C present receiver operating characteristic (ROC) curves and B and D present precision-recall curves (PRC). D (diagnosis), M (medication), A (age), and Y (year) represent the modalities being used for model training and evaluation.

## Encounter contribution results

Based on model predictive performance, we formed Dataset C for contribution analysis with patients that fall into the first and fifth AUROC quintile (Figure S3).

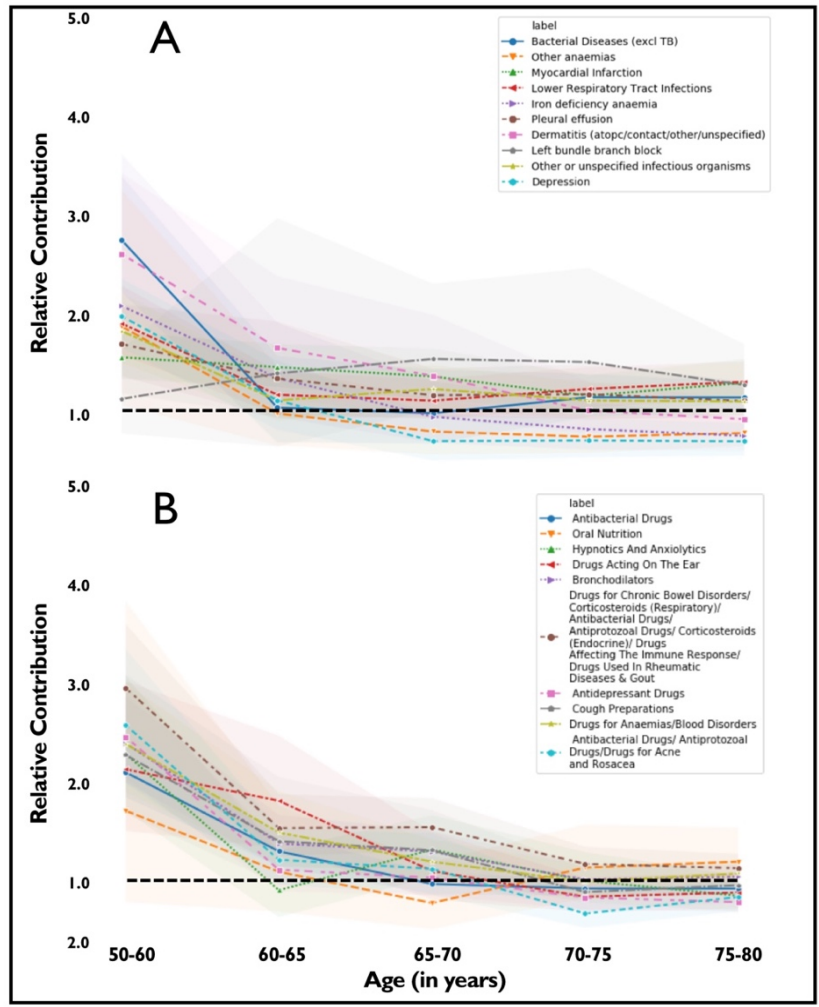

**Figure 3: Age-stratified relative contribution analyses for diseases and medications identified by model.** A and B represent top 10 relative contributions of diseases and medications to heart failure prediction respectively. X and y axis represent age groups in year and relative contribution (mean and shaded 95% confidence interval), respectively. Relative contribution equals to 1 implies equal contribution to both heart failure and non-heart failure predictions. Figures show that the relative contribution of risk factors to heart failure prediction tends to attenuate in older age. The strongest contributors in the medication category tend to have a corresponding disease, supporting the relatedness of the medication-disease pairs. The black dotted line denotes 1.0 relative contribution.

Tables S2 and S3 show RC for diseases and medications, respectively. We found diseases like bacterial diseases, lower respiratory tract infections, myocardial infarction, and pleural effusion and medications such, as "corticosteroid /antibacterial drugs", "bronchodilators", and "acne and rosacea drugs" were all positively associated with HF. Furthermore, our age-stratified RC analysis for the ten most important (i.e. highest contribution) diagnoses (Figure 3A) and medications (Figure 3B) showed the contributions were generally higher for those aged 50-60 years and lower in older ages implying little contribution of the risk factors individually to HF in older ages (consistent with evidence from epidemiological studies). For some predictors (e.g. left bundle branch block), however, CIs were too wide to allow firm conclusions about any differential RC by age (Tables S4 and S5).

Additionally, many of the medications showed a high RC to HF (Figure 3B) are treatments for diagnoses (Figure 3A). This implies the model identified diagnoses and treatments that were at least contemporaneous and often causally associated. For example, dermatitis may be treated with corticosteroids, which may be linked to cardiovascular risk[18] as may be depression, and therefore its treatments[19]. And "cough preparations", "bronchodilators" and "antibacterial drugs" are often linked to lower respiratory tract infection, asthma and chronic obstructive lung disease.[20,21] This could signal delayed diagnosis of HF due to misattribution of HF symptoms to respiratory diseases[21]. Or it might be that direct effects of drugs such as non-steroidal anti-inflammatory drugs (NSAIDs) are at least in part responsible for causing HF.[22]

Moreover, we captured RC for established HF risk factors (Table S6). Ischemic stroke, myocardial infarction, diabetes, hypertension, and atrial fibrillation/flutter all had RC>1. The age-stratified RC for these diseases (Figure 4/ Table S7) showed a similar pattern to the top ten diagnoses and medications analysis, again implying limited discriminatory contribution of these predictors individually in older people.

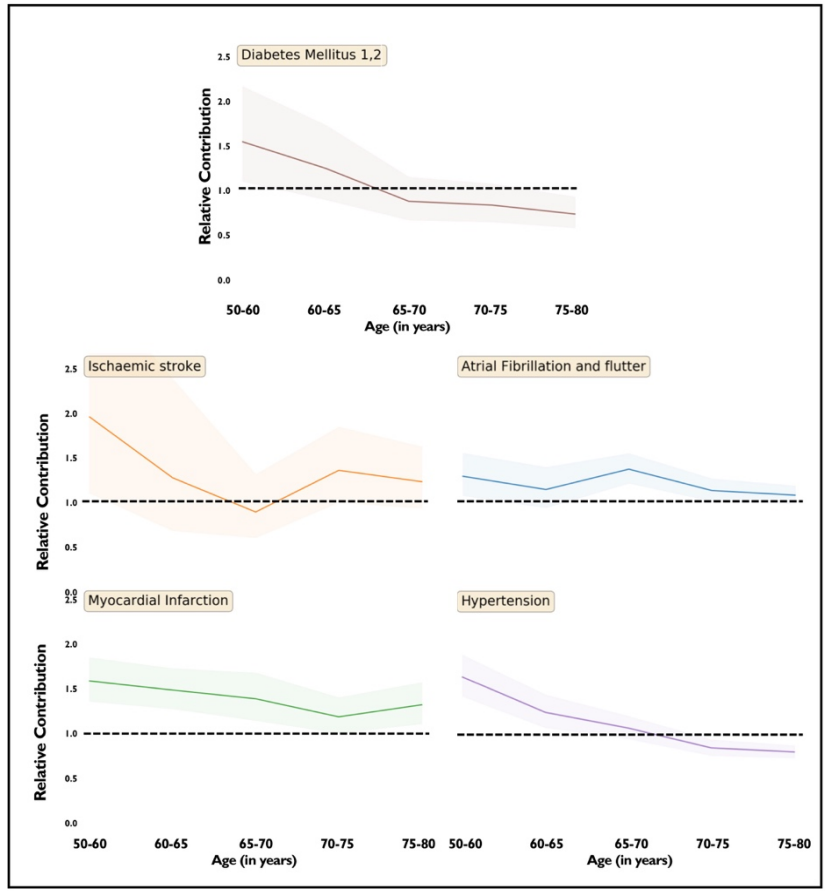

**Figure 4: Age-stratified relative contribution analyses for established risk factors**. X and y axis represent age groups in year and relative contribution. Relative contribution greater than 1 implies more contribution to heart failure prediction while less than 1 implies more contribution to the non-heart failure prediction. Each relative contribution is presented with a mean and 95% confidence interval. The black dotted line denotes 1.0 relative contribution.

We further explored factors that were strongly associated with non-HF (lowest RC values) with RC <1. To illustrate, Figure 5A and Table S5 show that treatments for established risk factors for HF, such as hypertension, diabetes and atrial fibrillation, including digoxin, were associated with a lower risk of HF.[23]

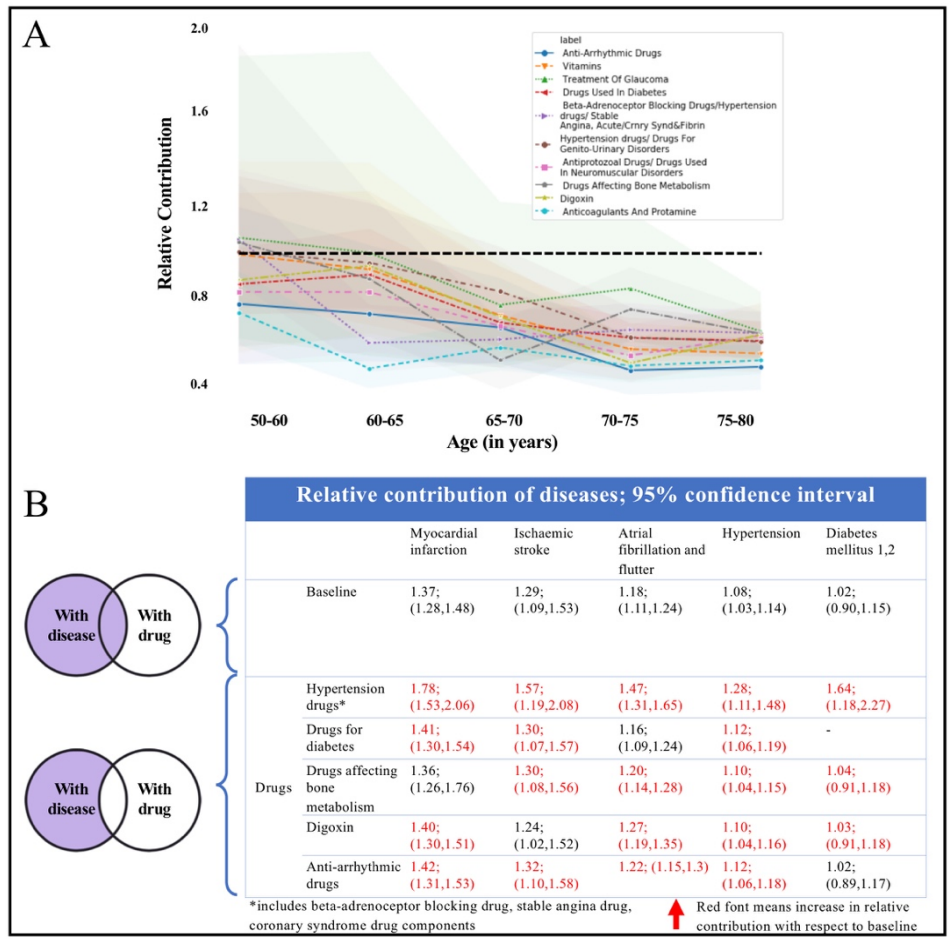

**Figure 5: Lowest age-stratified relative contributions for medications and contextuality of medications and diagnoses analysis.** A, Bottom 10 relative contributions of medications to heart failure prediction. X and y axis represent age groups in year and relative contribution (mean and 95% confidence interval). The black dotted line denotes 1.0 relative contribution. B, relative contribution (mean; 95% confidence interval) for established risk factors stratified by prescription status. The first row (baseline) shows relative contribution of all patients with the disease denoted by column (left Venn circle shaded). Subsequent rows describe relative contribution calculated on the subpopulation of patients who have not taken the drug denoted in a particular row (left Venn subsection of circle shaded). The red color denotes increase in relative contribution with respect to baseline disease relative contribution. "-" denotes no measurement due to insufficient size in those subgroups (too few patients with diabetes not receiving hypoglycemic therapy).

To disentangle the relationships between disease and medication pairs, we repeated the analyses only for patients who were not treated for a particular established risk factor. We show in Figure 5B that in untreated subgroups, there is a general increase in the RC for each disease (20 of 25 cases). This change in RC in people with or without treatment might also explain some of the unexpected patterns observed. For instance, although the overall and age-specific patterns of RC scores of established HF risk factors roughly concur with pre-existing epidemiological evidence, hypertension is shown to have <1 RC in ages, 70-80 years in Figure 4 (suggesting that people diagnosed with hypertension in older age are at lower risk of HF, whereas the opposite is found in younger people). Upon further investigation, we see that 74.5% of patients with hypertension were treated with "hypertension drugs," which had lowest RC values (Table S5). Repeating the analysis in people who were not treated for hypertension, we indeed find no evidence of a 'protective' effect of hypertension in older age groups; rather, the general trend of RC in older ages converging to 1.0 is preserved and are shown to be: 1.06; 95% CI (0.82,1.37) and 0.98; 95% CI (0.78, 1.23) for the 70-75 and 75-80 age group respectively.

Lastly, we stratified by calendar year-groups to investigate historical trends in treatments in Figure 6A. In Figure 6B, between 1990 and 2010, we show the number of times a medication was first prescribed in patients from Dataset A (not counting repeat prescriptions).

Throughout the 1990's, timolol, a beta blocker, was a common topical treatment for glaucoma[24] but with known cardiovascular side-effects such as bradycardia with potential to exacerbate HF[25]. With the introduction of new medications in the 2000's, the use of ophthalmic timolol started to decline[26] (Figure 6B). BEHRT captures this change over time in Figure 6A: treatments for glaucoma prior to 2000 had a high RC to HF while forms of treatment after 2000 had little contribution. Specifically following 2000, BEHRT identifies that treatment for glaucoma – namely, prostaglandin analogues – has <1 RC, indicating potentially protective effects on HF incidence. Prostaglandins and analogues such as prostaglandin $I_2$ [27,28] and others[29] have vasodilating properties with the potential of reducing cardiovascular risk, although large-scale randomized trials to investigate preventative effects of this vasodilator are currently lacking.[27,28] Also, for digoxin, we note that prescription of digoxin wanes following 2005. However, RC for this positive inotropic drug remains stable in any year strata, thus further lending support to the hypothesis that digoxin could play a role in prevention of HF. Discussion of calendar year temporal RC trends of analgesics are provided in appendix section 7.

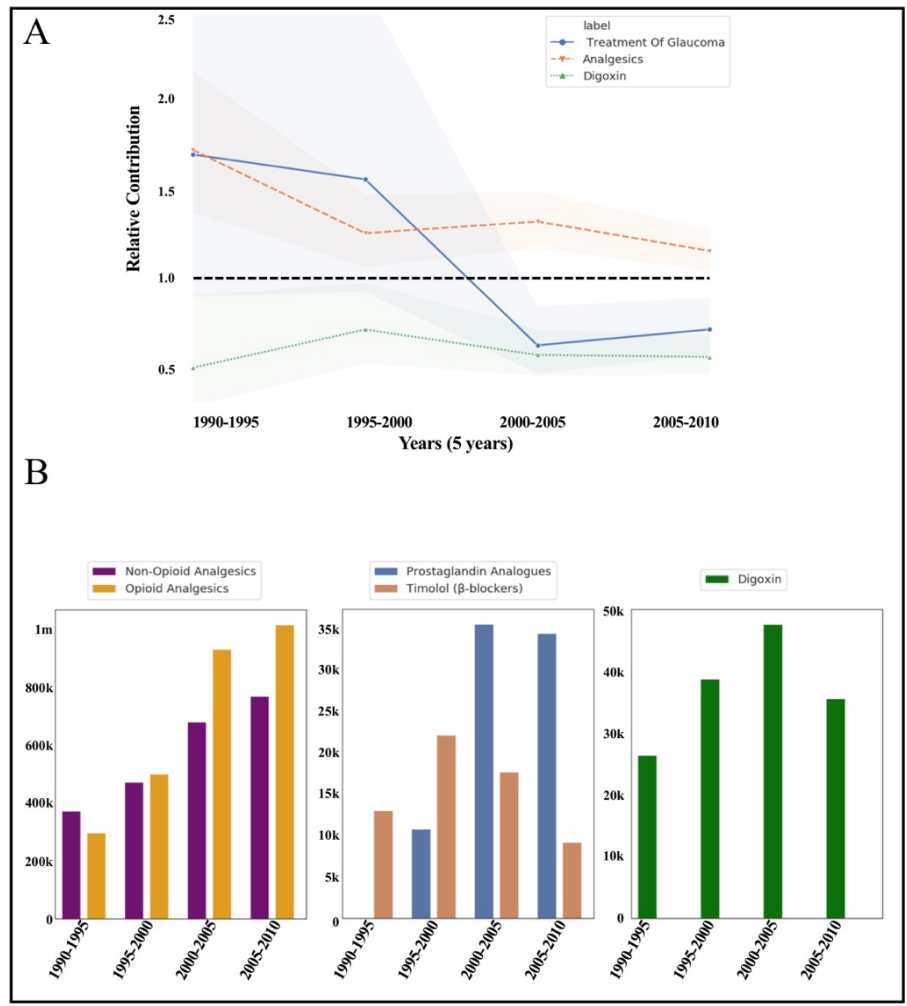

**Figure 6: Year-stratified relative contribution of medications.** A, relative contributions (mean and shaded 95% confidence interval) of three medications to heart failure prediction stratified by year. X and y represent year group and relative contribution respectively; a line is included to denote the 1.0 relative contribution. B, frequency of drugs by component in different year group in Dataset A. X and y represent the year group and counts of first-time prescriptions of drug component to patients respectively. Individual drug components are represented with bars in different color.

## Discussion

In this study, by leveraging large-scale routine EHR and machine learning, we developed and validated a model for predicting incident HF with superior performance compared to state-of-the-art, RETAINEX. Contextualization of clinical information is a major strength of the model and allows for flexible incorporation of other EHR modalities. We investigated the contribution and importance of various modalities and found that diseases and medications were strong predictors. Also, compared to age, calendar year improved patient representation substantially. We further developed an explainable framework for discovering factors contributing to risk of HF. This confirmed the relative importance of several established risk factors[15] and provided insights into medications that might negatively or positively contribute to the HF prediction.

Our work has several novel discoveries and methodological improvements. Our expansion of perturbation-based techniques[13] provides a method for making deep learning models explainable with findings supporting the importance of context in prediction. The RC method identified several disease-medication pairs, with one or both members of the pair being potential risk factors for cardiovascular diseases and specifically, HF in some cases. Also, we saw that our findings were broadly consistent with prior knowledge of HF and corroborated disease and medication risk factors of the disease.

With this approach, we included many potentially predictive variables not previously included in epidemiological studies. In addition, although age is usually incorporated as a risk factor for risk prediction,[6,12] our analysis found that incorporating calendar year provided additional and stronger information for accurate prediction of incident HF. A potential explanation for this observation was provided in our perturbation analysis stratified by year. This showed that occurrence of medications over different years made quantitatively different contributions to disease prediction, which would be missed if temporal context was not included in the models. Changes in such predictors over time or more subtle changes in disease patterns for instance due to advances in technologies leading to more accurate and frequent diagnosis are well known to clinicians. BEHRT enables incorporation of such information for better prediction. With regards to disease-medication contexts, a cursory analysis might lead to false conclusions that for instance treatments for hypertension increase the risk of HF. However, this conclusion is biased by indication; the correct interpretation is that the medication serves as a proxy for hypertension, which has a strong effect on HF incidence. Overall, the RC analysis illuminates potential protective medications that warrants further analysis in future studies.

Additionally, through age and year stratified analysis, BEHRT demonstrates medications with potentially preventative effects on HF. In the case of digoxin and prostaglandin analogues, the stable RC <1 in both age and year stratification signals potentially preventative effects of these drugs. However, as with standard statistical models, a causal interpretation should be made with great caution. Rather, our method provides a way of making models explainable and generates hypotheses, which depending on the totality of evidence from this work and other sources, should provide the impetus for additional confirmatory studies.

## Study Limitations

Our study has some additional limitations. The phenotyping method for diagnoses maps codes to 299 disease categories[30] losing information in the original granularity of the disease encoding and potentially biased by an expert's preferences. Also, the current work used limited information available within EHR and validated the model prediction performance on one dataset, CPRD. Future studies could explore whether other records, such as measurements, blood tests, and other demographics information (e.g. ethnicity, sex) will help improve model accuracy and explainability as well as model transferability to other datasets. Additionally, during the cohort selection, we kept patients with sufficient records to make robust predictions. This can potentially compromise model's generalizability for prediction in low risk groups who have fewer clinical encounters.

We developed a superior model for prediction of incident HF using routine EHR data providing a promising avenue for research into prediction of other complex conditions. Incorporating BEHRT into routine EHR could alert clinicians to those at risk for more targeted preventive care or recruitment into clinical trials. In addition, we highlight a data-driven approach for identification of potential risk factors that generate new hypotheses requiring causal exploration. We note that there are several medications which contribute negatively to the HF prediction. Not only are many of them used to treat established risk factors of HF, but others have not been tested for such an indication and might provide a starting point for drug repurposing studies. The model and analysis could be applied to more deeply phenotyped populations for discovery of new disease mechanisms and patterns in other complex conditions.

## Sources of funding

This work was supported by British Heart Foundation [PG/18/65/33872] (United Kingdom), United Kingdom Research and Innovation Global Challenges Research Fund [ES/P0110551/1] (United Kingdom), National Institute for Health Research (United Kingdom), Oxford Biomedical Research Centre (Oxford, United Kingdom), Oxford Martin School (Oxford, United Kingdom)

## Disclosures

Relationships with Industry: None to report.

Disclosures: All authors have nothing to disclose.

# Central Illustration (Figure 1)

## A Dataset A for pre-training dataset

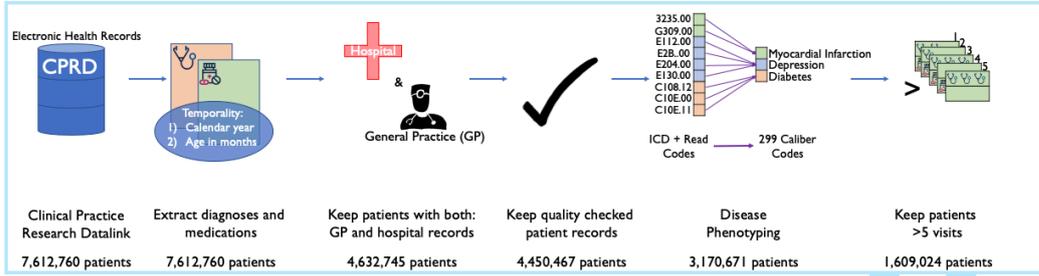

| Clinical Practice Research Datalink | Extract diagnoses and medications | Keep patients with both: GP and hospital records | Keep quality checked patient records | Disease Phenotyping | Keep patients >5 visits |
|---|---|---|---|---|---|
| 7,612,760 patients | 7,612,760 patients | 4,632,745 patients | 4,450,467 patients | 3,170,671 patients | 1,609,024 patients |

## B Dataset B for incident heart failure dataset

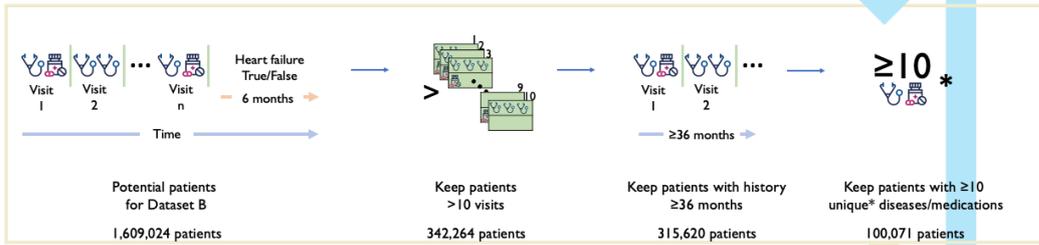

| Potential patients for Dataset B | Keep patients >10 visits | Keep patients with history ≥36 months | Keep patients with ≥10 unique* diseases/medications |
|---|---|---|---|
| 1,609,024 patients | 342,264 patients | 315,620 patients | 100,071 patients |

## C Data representation: BEHRT embeddings

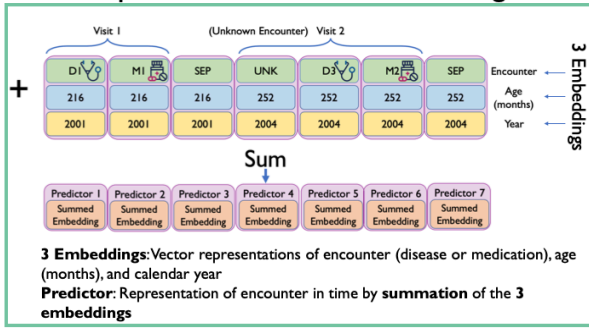

**3 Embeddings:** Vector representations of encounter (disease or medication), age (months), and calendar year
**Predictor:** Representation of encounter in time by **summation** of the 3 embeddings

## D Pre-training

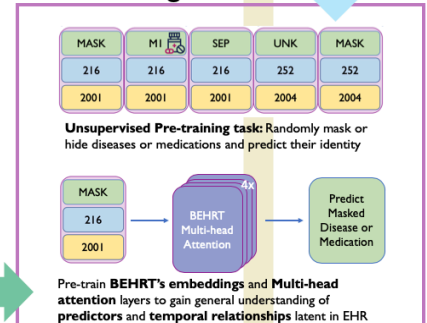

**Unsupervised Pre-training task:** Randomly mask or hide diseases or medications and predict their identity

Pre-train **BEHRT's embeddings** and **Multi-head attention** layers to gain general understanding of **predictors** and **temporal relationships** latent in EHR

## G Dataset C for contribution dataset

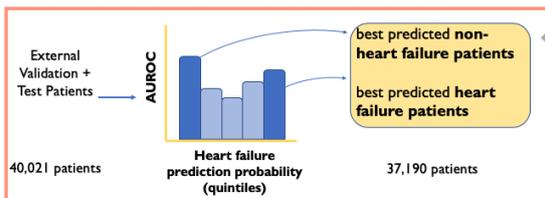

40,021 patients     37,190 patients

## E Heart failure task

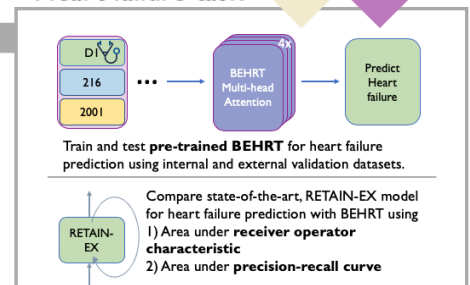

Train and test **pre-trained BEHRT** for heart failure prediction using internal and external validation datasets.

Compare state-of-the-art, RETAIN-EX model for heart failure prediction with BEHRT using
1) Area under **receiver operator characteristic**
2) Area under **precision-recall curve**

## H Contribution Analysis

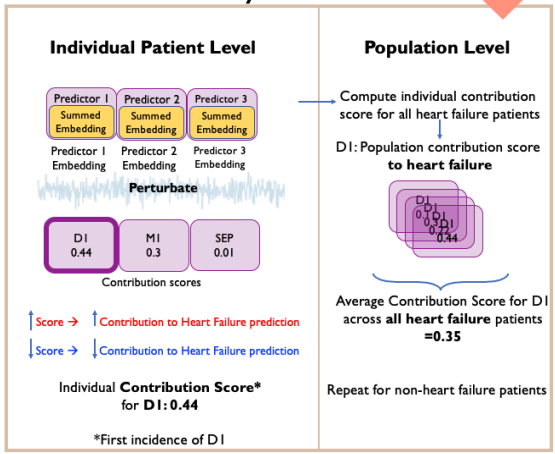

**Individual Patient Level** | **Population Level**

Compute individual contribution score for all heart failure patients

D1: Population contribution score **to heart failure**

Average Contribution Score for D1 across **all heart failure** patients = 0.35

Repeat for non-heart failure patients

↑ Score → ↑ Contribution to Heart Failure prediction
↓ Score → ↓ Contribution to Heart Failure prediction

Individual **Contribution Score*** for **D1: 0.44**

*First incidence of D1

## F Analysis: Modalities

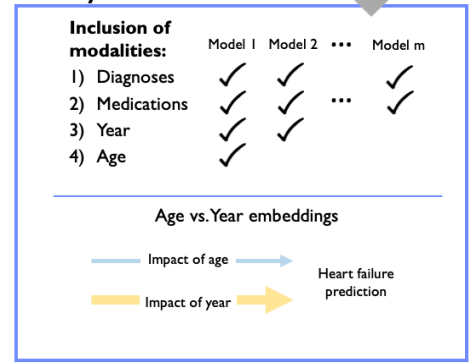

Inclusion of modalities:
1) Diagnoses
2) Medications
3) Year
4) Age

Age vs. Year embeddings

Impact of age → Heart failure prediction
Impact of year →

**Figure 1. Central Illustration.** (A, D) Dataset A is used for general contextual unsupervised pretraining. (B, E) With additional cohort selection criteria, we further design Dataset B for incident heart failure prediction. (C) Predictors are represented as summed embeddings. Codes not phenotyped are shown as "UNK"; "D#" and "M#" represent hypothetical diagnoses and medications respectively. "SEP" represents visit separation. (F) We assessed the utility of modalities for predictive performance. (G) We form Dataset C by selecting patients who are predicted most accurately (AUROC bins) for relative contribution analysis. (H) We develop population-based relative contribution by aggregating the individual-level predictor contributions.

# Supplementary Material

## 1. Details for BEHRT model

In NLP literature and BERT,[1] words in sentences are considered "tokens" and sentences are separated from one another with a separation element. Similarly, we conceptualized medical events such as diagnoses and medications in a doctor/hospital visit as encounters (or tokens) and separate visits by a separation element ("SEP"). Similar to the original BERT model, we implemented an annotation ordering the sequential medical history data. Furthermore, we added layers of information that involve age and calendar year of each encounter. Thus, the total input comprised of three layers of information for each and every encounter: the encounter itself (diagnoses and/or medications), age, and calendar year.

## 2. Visualised importance analysis for age and year

To further validate and visualise the contribution of two time-related modality (i.e. age and year), we conducted similarity measurement (cosine similarity[2]) across the summation of embeddings for encounters and calendar years/ages to observe if there is any significant difference for their representations. The larger the distance, the higher the dissimilarity, and the higher dissimilarity can directly imply the importance of a modality. Embedding here means a matrix that is trained to represent each element in a modality, such as hypertension in diagnose and 2011 in year.

Fig S2A and Fig S2B show two cosine similarity matrices for each pair of instances in embeddings of calendar year and age. We chose four diseases with high occurrence in the dataset to ensure these embeddings were well-trained. Calendar year showed substantial dissimilarity across years (from 0 to 0.8). By contrast, the dissimilarity as consequence of variation in age in months was less pronounced (from 0 to 0.3). In other words, representation of diseases (or predictors) among individuals was more sensitive to variation in year in which they were recorded than to the age of the patient. This suggests that calendar year or 'birth cohort effect' was more informative for the incident HF prediction than chronological age, as it captures more contextual time-sensitive information necessary for HF prediction.

## 3. Details of perturbation methods

Additionally to Guan et al.'s method[3], we developed an asymmetric loss function to prioritize learning perturbations in addition to the information entropy-based loss term. Shown in the equations below is the full description of the loss function conducted over one patient's medical history.

$$alpha(y, x', s) = \begin{cases} \beta_1, & \text{if } y = 1, \Phi(\tilde{x}) - s \geq 0 \\ \beta_1, & \text{if } y = 0, \Phi(\tilde{x}) - s \leq 0 \quad (1) \\ \beta_2, & \text{otherwise} \end{cases}$$

$$L(\sigma) = alpha(y, x', s) \times \mathbb{E}_\epsilon \|(\Phi(x') - s)\|^2 - \lambda \sum_{i=1}^{n} H(X'_l | s)\big|_{\epsilon_i \sim \mathcal{N}(0, \sigma_i^2 I)} \quad (2)$$

$\tilde{x}$: perturbed input encounter embeddings (original represented by $x$)

$n$: number of encounters in this patient's medical history

$s$: output state of original input (without perturbation)

$\Phi(\tilde{x})$: output state of perturbed input

$\beta_1, \beta_2$: weight hyperparameters ($\beta_1 < \beta_2$); if $\beta_1 = \beta_2$, then loss function is symmetric.

$y$: heart failure label

$\mathbb{E}_\epsilon \|\Phi(\tilde{x}) - s\|^2$: mean squared error described in Guan et al.[3]

$alpha(y, \tilde{x}, s)$: asymmetric weight function

$\lambda \sum_{i=1}^{n} H(\tilde{X}_l | s)\big|_{\epsilon_i \sim \mathcal{N}(0, \sigma_i^2 I)}$: information entropy based loss function described in Guan et al.[3]

    To describe Eq. (1) and (2) in words, for heart failure patients, we prioritize perturbations that increases the outcome probability than those that decreases it; and we prioritize the opposite for non-heart failure patients. To do so, we penalize with the *alpha(y,x',s)* constant. Asymmetric losses are often used in scenarios where an error in one direction (perhaps positive) is more costly than an error in the opposite direction.[4,5] This perturbation method delivers learned $\epsilon = [\epsilon_1, \epsilon_2, ..., \epsilon_n]$ with $\epsilon_I$ per predictor $i$ – the trained, allowable variance for the predictor, with maximum variance defined by a user defined hyperparameter (set to 0.5 in our work). To assess contribution of predictor, we transform $\epsilon_I$ to 0.5- $\epsilon_I$ to reflect the inverse relationship: the lower the $\epsilon_I$, the higher contribution to heart failure prediction and vice-versa. As seen in Fig 1H, we first establish patient-level contribution, or 0.5- $\epsilon_I$ for a particular encounter (disease/medication) in time.

## 4. Discussion of Dataset C

The encounter contribution analysis was conducted on a subset of patients from the external testing and validation datasets where BEHRT (data modalities: DMAY) performed best named Dataset C. Since BEHRT has learnt the latent space most effectively for these patients, the resulting contribution analysis on these patient medical histories are therefore reliable. We avoided selecting incorrectly predicted patients for this analysis; even if the perturbation method assigns theoretically correct derivations of predictor contribution to output probability, when conducting population-wise analysis and age-stratified analysis as described, misclassification of the outcome would invalidate results for this particular analysis. Thus, the data from these best-performing groups were selected for contribution analysis.

5. Details of disease and medication phenotyping

In our study, we used all diagnostic codes and medications at each encounter as well as patient age in months and calendar year. Encounters represent each individual's time-stamped recording of a diagnosis or medication. For medication specifically, we included all available prescription records as new encounters in the dataset. Because of how CPRD records medications in six-week increments, medications make up the largest number of encounters.

In the UK, primary care and secondary care use different coding systems for diagnosis (i.e. Read Code[6] in GP and 10th revision of International Statistical Classification of Diseases and Related Health Problems [ICD-10] Code[7] in Hospital Episode Statistics). We mapped these codes using a dictionary given by CPRD. This led to 56,624 unique codes, which were then mapped to 299 clinically meaningful disease categories, using Caliber, a previously published and clinically-validated phenotyping method.[8] Medication codes were classified using the British National Formulary hierarchical coding format.[9] We used codes at the section level prevalent in the population leading to 426 unique medication group codes. Both disease and medication codes with unknown mapping was mapped to an "UNKNOWN" category. The "heart failure" phenotype is defined by Caliber to be a collection of Read and ICD-10 codes. We looked at the diagnoses codes strictly (as opposed to codes of historical diagnoses). The codes are found at: https://www.caliberresearch.org/portal/phenotypes/heartfailure, using incident codes from primary (Read) and secondary care (ICD-10).

6. Details of model training

The model was implemented using PyTorch[10]. We applied Bayesian optimization[11] for model hyperparameter tuning on the number of layers, hidden size and intermediate size. After 20 iterations of searching the parameter space, we chose the optimal hyperparameters for the model, with number of layers: 4, hidden size: 120, number of attention heads: 6, and intermediate size: 108. We pre-trained BEHRT's weights using the Masked Language Modelling[1] pre-training task using the dataset A: we randomly masked some encounters in the medical history of the patient, and predicted the masked encounters. This task is unsupervised and undertaken to let the model gain a general understanding of the predictors and their temporal relationship in the longitudinal data. After pre-training the model, we implemented the model for the heart failure prediction task on dataset B.

## 7. Temporal trends of analgesics

For, analgesics, we see a similar declining trend as calendar years increase. In Fig 6B, we first show that prescription of analgesics was more non-opioid than opioid prior to 1996. We see that BEHRT parallels this generational shift in drug component prescription through RC; non-opioid analgesics are primarily composed of NSAID's and generally increase the risk of cardiovascular events[12,13]. Thus, RC prior to 1995 is shown to be quite high. Tracing the gradual change in majority preference to favour opioid based analgesics following 1995, RC captures this change in prescription behaviour and as a result, attenuates in the following decades.

## 8. Additional results

**S1 Table: Characteristics of patients in the training, test, and validation set**

|  | Training Set | Test Set | Validation Set |
| --- | --- | --- | --- |
| Total number of patients (%) | 60,043 (100) | 20,014 (100) | 20,014 (100) |
| Number of incident cases of heart failure (%) | 7,853 (13.1) | 2,621 (13.1) | 2,576 (12.9) |
| Women (%) | 35,094 (58.4) | 11,634 (58.1) | 11,603 (58.0) |
| Men (%) | 24,949 (41.6) | 8,380 (41.9) | 8,411 (42.0) |
| Median follow-up duration (year) | 9 | 9 | 10 |
| Median age (year) | 70 | 70 | 70 |
| Diabetes Mellitus (%) | 12,348 (20.5) | 4,130 (20.6) | 4,128 (20.6) |
| Hypertension (%) | 39,427 (65.6) | 13,096 (65.4) | 13,237 (66.1) |
| Rheumatoid arthritis (%) | 1,936 (3.2) | 673 (3.4) | 679 (3.4) |
| Atrial fibrillation and flutter (%) | 15,826 (26.3) | 5,179 (25.9) | 5,252 (26.2) |
| Myocardial infarction (%) | 5,588 (9.3) | 1,851 (9.2) | 1,839 (9.2) |
| Chronic obstructive pulmonary disease (COPD) (%) | 8,364 (13.9) | 2,770 (13.8) | 2,763 (13.8) |
| Ischemic stroke (%) | 3,022 (5.0) | 1,037 (5.2) | 1,065 (5.3) |

**S2 Table: Relative contribution scores for diseases that occurred in at least 5% of the population.**

| Disease | HF | Standard Deviation: HF | Non-HF | Standard Deviation: Non-HF | RC | CI Lower bound | CI Upper bound |
|---|---|---|---|---|---|---|---|
| Dermatitis (atopic/contact/other/unspecified) | 0.12 | 0.1 | 0.07 | 0.08 | 1.76 | 1.56 | 1.98 |
| Bacterial diseases (excluding tuberculosis) | 0.23 | 0.13 | 0.15 | 0.13 | 1.56 | 1.42 | 1.71 |
| Other or unspecified infectious organisms | 0.24 | 0.13 | 0.16 | 0.13 | 1.47 | 1.37 | 1.58 |
| Lower respiratory tract infections | 0.28 | 0.13 | 0.19 | 0.13 | 1.47 | 1.37 | 1.57 |
| Depression | 0.13 | 0.11 | 0.09 | 0.1 | 1.45 | 1.29 | 1.63 |
| Iron deficiency anaemia | 0.15 | 0.12 | 0.11 | 0.11 | 1.4 | 1.22 | 1.61 |
| Pleural effusion | 0.33 | 0.12 | 0.24 | 0.13 | 1.38 | 1.24 | 1.53 |
| Myocardial infarction | 0.32 | 0.12 | 0.24 | 0.13 | 1.37 | 1.28 | 1.48 |
| Left bundle branch block | 0.33 | 0.12 | 0.25 | 0.12 | 1.34 | 1.13 | 1.60 |
| Other anaemias | 0.17 | 0.13 | 0.13 | 0.13 | 1.33 | 1.17 | 1.50 |
| Hypo or hyperthyroidism | 0.17 | 0.12 | 0.13 | 0.12 | 1.26 | 1.13 | 1.41 |
| Atrial fibrillation and flutter | 0.23 | 0.14 | 0.20 | 0.13 | 1.18 | 1.11 | 1.24 |
| Acute kidney injury | 0.27 | 0.13 | 0.23 | 0.14 | 1.15 | 1.01 | 1.31 |
| Hypertension | 0.13 | 0.10 | 0.12 | 0.11 | 1.08 | 1.03 | 1.14 |
| Hearing loss | 0.12 | 0.11 | 0.12 | 0.11 | 1.03 | 0.86 | 1.22 |
| Urinary tract infections | 0.15 | 0.11 | 0.15 | 0.13 | 1.00 | 0.88 | 1.12 |
| Enthesopathies & synovial disorders | 0.08 | 0.07 | 0.09 | 0.10 | 0.9 | 0.79 | 1.02 |
| Stroke not otherwise specified (NOS) | 0.17 | 0.10 | 0.19 | 0.12 | 0.89 | 0.79 | 0.99 |
| Chronic obstructive pulmonary disease (COPD) | 0.17 | 0.12 | 0.19 | 0.13 | 0.87 | 0.80 | 0.95 |
| Cataract | 0.12 | 0.10 | 0.14 | 0.11 | 0.84 | 0.75 | 0.94 |
| Hyperplasia of prostate | 0.11 | 0.09 | 0.13 | 0.12 | 0.82 | 0.72 | 0.95 |
| Stable angina | 0.16 | 0.12 | 0.19 | 0.13 | 0.81 | 0.75 | 0.88 |
| Osteoarthritis (excluding spine) | 0.10 | 0.09 | 0.13 | 0.12 | 0.75 | 0.68 | 0.84 |
| Diabetic ophthalmic complications | 0.13 | 0.11 | 0.18 | 0.12 | 0.7 | 0.62 | 0.79 |

**In the columns, we have heart failure (HF) patients average contribution scores, non-heart failure (Non-HF) patients average contribution scores, relevant standard deviation measures, and relative contribution (RC) and 95% confidence interval (CI) for diseases.**

**S3 Table: Relative contribution scores for medications that occurred in at least 5% of the population.**

| Medication | HF | Standard Deviation: HF | Non-HF | Standard Deviation: Non-HF | RC | CI LB | CI UB |
|---|---|---|---|---|---|---|---|
| Chronic bowel disorders/ corticosteroids (respiratory)/ antibacterial drugs/ antiprotozoal drugs/ corticosteroids (endocrine)/ drugs affecting the immune response/ drugs used in rheumatic diseases & gout | 0.14 | 0.11 | 0.08 | 0.08 | 1.71 | 1.59 | 1.84 |
| Anaemias and other blood disorders | 0.09 | 0.08 | 0.06 | 0.07 | 1.53 | 1.42 | 1.66 |
| Bronchodilators | 0.11 | 0.10 | 0.08 | 0.08 | 1.52 | 1.43 | 1.63 |
| Cough preparations | 0.14 | 0.11 | 0.10 | 0.10 | 1.45 | 1.28 | 1.64 |
| Oral nutrition | 0.12 | 0.10 | 0.08 | 0.08 | 1.41 | 1.68 | 1.68 |
| Antibacterial drugs/ antiprotozoal drugs/ acne and rosacea | 0.11 | 0.10 | 0.08 | 0.09 | 1.41 | 1.25 | 1.58 |
| Hypnotics and anxiolytics | 0.09 | 0.09 | 0.07 | 0.07 | 1.39 | 1.22 | 1.57 |
| Antidepressant drugs | 0.08 | 0.07 | 0.06 | 0.07 | 1.38 | 1.28 | 1.50 |
| Drugs acting on the ear | 0.11 | 0.09 | 0.08 | 0.09 | 1.36 | 1.2 | 1.54 |
| Antibacterial drugs | 0.09 | 0.08 | 0.07 | 0.07 | 1.32 | 1.26 | 1.39 |
| Antibacterial drugs/ acne and rosacea | 0.09 | 0.08 | 0.07 | 0.08 | 1.31 | 1.2 | 1.44 |
| Drugs used in nausea and vertigo | 0.11 | 0.09 | 0.08 | 0.09 | 1.27 | 1.16 | 1.40 |
| Soft-tissue disorders and topical pain relif | 0.08 | 0.07 | 0.06 | 0.07 | 1.24 | 1.13 | 1.36 |
| Drugs acting on the oropharynx | 0.10 | 0.08 | 0.08 | 0.08 | 1.22 | 1.07 | 1.41 |
| Hypnotics and anxiolytics/ general anaesthesia | 0.09 | 0.09 | 0.08 | 0.08 | 1.22 | 1.06 | 1.40 |
| Analgesics/ analgesics | 0.07 | 0.06 | 0.06 | 0.06 | 1.21 | 1.13 | 1.29 |
| Corticosteroids (respiratory) | 0.08 | 0.07 | 0.07 | 0.07 | 1.20 | 1.09 | 1.32 |
| Drugs used in rheumatic diseases and gout | 0.07 | 0.07 | 0.06 | 0.07 | 1.19 | 1.11 | 1.28 |
| Anti-infective eye preparations | 0.09 | 0.08 | 0.07 | 0.08 | 1.17 | 1.05 | 1.05 |
| Laxatives | 0.07 | 0.07 | 0.06 | 0.07 | 1.16 | 1.08 | 1.25 |
| Anti-infective skin preparations | 0.08 | 0.07 | 0.07 | 0.08 | 1.14 | 1.01 | 1.29 |
| Vaccines and antisera | 0.11 | 0.08 | 0.09 | 0.09 | 1.13 | 1.06 | 1.20 |
| Antihistamine, hyposensitivity and allergic emergent | 0.07 | 0.06 | 0.07 | 0.08 | 1.11 | 1.01 | 1.22 |
| Dyspepsia and gastro-oesophageal reflux disease | 0.08 | 0.07 | 0.08 | 0.08 | 1.09 | 0.98 | 1.20 |
| Antiplatelet drugs | 0.10 | 0.09 | 0.09 | 0.09 | 1.08 | 1.02 | 1.14 |
| Emollient and barrier preparations | 0.08 | 0.07 | 0.07 | 0.07 | 1.08 | 1.00 | 1.16 |
| Sex hormones | 0.07 | 0.06 | 0.07 | 0.08 | 1.07 | 0.92 | 1.25 |
| Nitrates, calcium-channel blocker and other antianginal drugs | 0.10 | 0.09 | 0.09 | 0.09 | 1.06 | 1.01 | 1.12 |

| Category | | | | | | | |
|---|---|---|---|---|---|---|---|
| Hypnotics and anxiolytics/ antiepileptic drugs/ antiepileptic drugs/ drugs used in neuromuscular disorders/ general anaesthesia | 0.08 | 0.06 | 0.07 | 0.08 | 1.06 | 0.94 | 1.20 |
| Miscellaneous ophthalmic preparations | 0.08 | 0.07 | 0.07 | 0.07 | 1.05 | 0.92 | 1.19 |
| Topical corticosteroids | 0.07 | 0.06 | 0.07 | 0.07 | 1.05 | 0.98 | 1.11 |
| Local preparations for anal and rectal disorders | 0.08 | 0.06 | 0.07 | 0.09 | 1.03 | 0.9 | 1.17 |
| Corticosteroids and other anti-inflammatory preparations | 0.08 | 0.07 | 0.08 | 0.09 | 1.02 | 0.88 | 1.17 |
| Drugs acting on the nose | 0.07 | 0.06 | 0.07 | 0.08 | 1.01 | 0.93 | 1.11 |
| Acute diarrhoea | 0.10 | 0.08 | 0.09 | 0.10 | 1.01 | 0.88 | 1.15 |
| Topical corticosteroids/ anti-infective skin preparations | 0.08 | 0.07 | 0.09 | 0.09 | 0.99 | 0.86 | 1.13 |
| Analgesics | 0.07 | 0.06 | 0.07 | 0.08 | 0.98 | 0.93 | 1.03 |
| Hypertension and heart failure | 0.10 | 0.09 | 0.10 | 0.10 | 0.95 | 0.91 | 1.00 |
| Drugs used in psychoses and Related disorders/ drugs used in nausea and vertigo | 0.08 | 0.07 | 0.08 | 0.09 | 0.95 | 0.82 | 1.11 |
| Diuretics | 0.15 | 0.13 | 0.16 | 0.13 | 0.93 | 0.88 | 0.98 |
| Acute diarrhoea/ cough preparations/ analgesics | 0.08 | 0.07 | 0.09 | 0.09 | 0.92 | 0.82 | 1.03 |
| Bronchodilators/ corticosteroids (respiratory) | 0.08 | 0.07 | 0.09 | 0.08 | 0.92 | 0.83 | 1.01 |
| Lipid-regulating drugs | 0.07 | 0.07 | 0.08 | 0.07 | 0.91 | 0.87 | 0.97 |
| Analgesics/ drugs used in rheumatic diseases and gout | 0.06 | 0.06 | 0.07 | 0.08 | 0.91 | 0.79 | 1.05 |
| Antidepressant drugs/ analgesics/ analgesics | 0.07 | 0.06 | 0.08 | 0.08 | 0.91 | 0.82 | 1.01 |
| Beta-adrenoceptor blocking drugs | 0.08 | 0.08 | 0.10 | 0.09 | 0.87 | 0.79 | 0.95 |
| Antisecretory drugs and mucosal protectants | 0.07 | 0.06 | 0.08 | 0.09 | 0.85 | 0.81 | 0.90 |
| Diuretics/ minerals | 0.07 | 0.06 | 0.08 | 0.08 | 0.85 | 0.77 | 0.93 |
| Drugs for Genito-urinary disorders | 0.07 | 0.06 | 0.08 | 0.08 | 0.81 | 0.73 | 0.91 |
| Thyroid and antithyroid drugs | 0.07 | 0.06 | 0.09 | 0.09 | 0.81 | 0.73 | 0.9 |
| Treatment of glaucoma | 0.07 | 0.07 | 0.09 | 0.09 | 0.77 | 0.64 | 0.92 |
| Hypertension and heart failure/ drugs for genito-urinary disorders | 0.07 | 0.07 | 0.10 | 0.09 | 0.71 | 0.65 | 0.78 |
| Drugs used in diabetes | 0.07 | 0.06 | 0.09 | 0.09 | 0.71 | 0.66 | 0.77 |
| Vitamins | 0.06 | 0.05 | 0.09 | 0.09 | 0.7 | 0.63 | 0.79 |
| Drugs affecting bone metabolism | 0.07 | 0.06 | 0.11 | 0.09 | 0.7 | 0.78 | 0.78 |
| Antiprotozoal drugs/ drugs used in neuromuscular disorders | 0.07 | 0.07 | 0.11 | 0.09 | 0.67 | 0.58 | 0.77 |
| Beta-adrenoceptor blocking drugs/ hypertension and heart failure/ stable angina, acute/coronary syndrome and fibrin | 0.11 | 0.1 | 0.16 | 0.11 | 0.67 | 0.61 | 0.73 |
| Positive inotropic drugs | 0.08 | 0.08 | 0.13 | 0.09 | 0.61 | 0.54 | 0.69 |
| Anti-arrhythmic drugs | 0.07 | 0.06 | 0.12 | 0.09 | 0.56 | 0.48 | 0.65 |
| Anticoagulants and protamine | 0.07 | 0.07 | 0.14 | 0.10 | 0.53 | 0.48 | 0.57 |

In the columns, we have heart failure (HF) patients' average contribution scores, non-heart failure (Non-HF) patients' average contribution scores, relevant standard deviation measures, and relative contribution (RC) and 95% confidence interval (CI) for medications with lower bound (LB) and upper bound (UB).

**S4 Table: Age-stratified, relative contribution (RC) and 95% confidence intervals (CI) for diseases that occurred in at least 5% of the test and validation dataset. In the columns, we have RC and CI for five age categories with lower bound (LB) and upper bound (UB).**

| Disease | 50-60 | | | 60-65 | | | 65-70 | | | 70-75 | | | 75-80 | | |
|---|---|---|---|---|---|---|---|---|---|---|---|---|---|---|---|
| | RC | CI | | RC | CI | | RC | CI | | RC | CI | | RC | CI | |
| | | LB | UB | | LB | UB | | LB | UB | | LB | UB | | LB | UB |
| Acute kidney injury | 1.01 | 0.72 | 1.41 | 1.28 | 0.91 | 1.82 | 1.05 | 0.74 | 1.48 | 1.08 | 0.87 | 1.34 | 1 | 0.79 | 1.26 |
| Atrial fibrillation and flutter | 1.23 | 1.02 | 1.47 | 1.08 | 0.89 | 1.32 | 1.30 | 1.15 | 1.47 | 1.07 | 0.96 | 1.20 | 1.02 | 0.93 | 1.12 |
| Bacterial diseases (excluding tuberculosis) | 2.71 | 2.20 | 3.35 | 1.04 | 0.70 | 1.56 | 0.99 | 0.69 | 1.42 | 1.15 | 0.95 | 1.38 | 1.14 | 1 | 1.3 |
| Cataract | 1.38 | 0.86 | 2.21 | 0.64 | 0.43 | 0.95 | 0.86 | 0.64 | 1.15 | 0.81 | 0.64 | 1.03 | 0.72 | 0.61 | 0.83 |
| Chronic obstructive pulmonary disease (COPD) | 1.11 | 0.87 | 1.42 | 0.98 | 0.78 | 1.22 | 0.83 | 0.69 | 1.00 | 0.72 | 0.61 | 0.86 | 0.77 | 0.66 | 0.89 |
| Depression | 1.95 | 1.64 | 2.33 | 1.12 | 0.82 | 1.51 | 0.71 | 0.52 | 0.98 | 0.72 | 0.55 | 0.94 | 0.71 | 0.56 | 0.89 |
| Dermatitis (atopic/contact/other/unspecified) | 2.58 | 1.96 | 3.38 | 1.64 | 1.14 | 2.36 | 1.36 | 0.94 | 1.97 | 1.01 | 0.79 | 1.30 | 0.93 | 0.77 | 1.12 |
| Diabetic ophthalmic complications | 0.90 | 0.66 | 1.23 | 0.75 | 0.53 | 1.05 | 0.55 | 0.40 | 0.76 | 0.51 | 0.39 | 0.67 | 0.63 | 0.51 | 0.77 |
| Enthesopathies and synovial disorders | 1.94 | 1.45 | 2..9 | 0.90 | 0.59 | 1.39 | 0.70 | 0.53 | 0.92 | 0.41 | 0.32 | 0.52 | 0.4 | 0.33 | 0.49 |
| Hearing loss | 1.82 | 1.01 | 3.26 | 1.15 | 0.69 | 1.93 | 0.46 | 0.29 | 0.74 | 0.72 | 0.53 | 0.97 | 0.77 | 0.59 | 1 |
| Hyperplasia of prostate | 1.86 | 0.80 | 4.32 | 0.85 | 0.59 | 1.24 | 1.00 | 0.78 | 1.29 | 0.65 | 0.49 | 0.86 | 0.56 | 0.45 | 0.7 |
| Hypertension | 1.59 | 1.39 | 1.83 | 1.21 | 1.05 | 1.40 | 1.04 | 0.93 | 1.17 | 0.83 | 0.75 | 0.92 | 0.79 | 0.73 | 0.86 |
| Hypo or hyperthyroidism | 1.93 | 1.44 | 2.58 | 1.02 | 0.70 | 1.49 | 1.27 | 1.02 | 1.59 | 0.84 | 0.67 | 1.07 | 0.8 | 0.66 | 0.96 |
| Iron deficiency anaemia | 2.06 | 1.19 | 3.57 | 1.34 | 0.95 | 1.89 | 0.95 | 0.69 | 1.31 | 0.83 | 0.62 | 1.10 | 0.76 | 0.61 | 0.94 |
| Left bundle branch block | 1.13 | 0.79 | 1.62 | 1.39 | 0.65 | 2.93 | 1.53 | 1.03 | 2.28 | 1.50 | 0.92 | 2.44 | 1.27 | 0.96 | 1.67 |
| Lower respiratory tract infections | 1.88 | 1.57 | 2.26 | 1.17 | 0.97 | 1.41 | 1.11 | 0.93 | 1.33 | 1.23 | 1.04 | 1.45 | 1.3 | 1.13 | 1.5 |
| Myocardial infarction | 1.55 | 1.33 | 1.80 | 1.45 | 1.25 | 1.68 | 1.36 | 1.13 | 1.63 | 1.16 | 0.99 | 1.37 | 1.29 | 1.09 | 1.53 |
| Osteoarthritis (excluding spine) | 0.94 | 0.58 | 1.52 | 1.02 | 0.73 | 1.41 | 0.54 | 0.43 | 0.67 | 0.48 | 0.39 | 0.60 | 0.62 | 0.52 | 0.73 |

| | | | | | | | | | | | | | | |
|---|---|---|---|---|---|---|---|---|---|---|---|---|---|---|
| Other anaemias | 1.85 | 1.07 | 3.21 | 0.98 | 0.67 | 1.45 | 0.81 | 0.58 | 1.12 | 0.75 | 0.59 | 0.96 | 0.79 | 0.66 | 0.94 |
| Other or unspecified infectious organisms | 1.81 | 1.49 | 2.18 | 1.11 | 0.91 | 1.36 | 1.23 | 1.03 | 1.47 | 1.11 | 0.96 | 1.30 | 1.1 | 0.97 | 1.25 |
| Pleural effusion | 1.68 | 1.36 | 2.07 | 1.33 | 0.94 | 1.90 | 1.17 | 0.95 | 1.43 | 1.18 | 0.94 | 1.48 | 1.11 | 0.94 | 1.31 |
| Stable angina | 1.00 | 0.84 | 1.20 | 0.82 | 0.66 | 1.02 | 0.85 | 0.71 | 1.01 | 0.70 | 0.59 | 0.82 | 0.65 | 0.56 | 0.75 |
| Stroke not otherwise specified (NOS) | 1.34 | 0.92 | 1.97 | 0.86 | 0.60 | 1.25 | 0.77 | 0.57 | 1.03 | 0.85 | 0.67 | 1.07 | 0.75 | 0.63 | 0.89 |
| Urinary tract infections | 1.51 | 1.06 | 2.15 | 0.72 | 0.44 | 1.17 | 0.82 | 0.55 | 1.23 | 0.66 | 0.52 | 0.83 | 0.78 | 0.66 | 0.92 |

**S5 Table: Age-stratified, relative contribution (RC) and 95% confidence intervals (CI) for medications that occurred in at least 5% of the test and validation dataset.**

**In the columns, we have RC and CI for five age categories with lower bound (LB) and upper bound (UB).**

| Medications | 50-60 | | | 60-65 | | | 65-70 | | | 70-75 | | | 75-80 | | |
|---|---|---|---|---|---|---|---|---|---|---|---|---|---|---|---|
| | RC | CI | | RC | CI | | RC | CI | | RC | CI | | RC | CI | |
| | | LB | UB | | LB | UB | | LB | UB | | LB | UB | | LB | UB |
| Acute diarrhoea | 1.79 | 1.24 | 2.58 | 1.2 | 0.87 | 1.65 | 0.79 | 0.58 | 1.1 | 0.68 | 0.51 | 0.91 | 0.53 | 0.43 | 0.66 |
| Acute diarrhoea/ cough preparations/ analgesics | 1.49 | 1.16 | 1.92 | 0.91 | 0.67 | 1.23 | 0.52 | 0.36 | 0.75 | 0.67 | 0.53 | 0.85 | 0.66 | 0.54 | 0.8 |
| Anaemias and other blood disorders | 2.36 | 1.87 | 2.99 | 1.45 | 1.12 | 1.88 | 1.15 | 0.96 | 1.38 | 0.94 | 0.8 | 1.11 | 1.03 | 0.9 | 1.17 |
| Analgesics | 1.62 | 1.37 | 1.9 | 0.93 | 0.79 | 1.09 | 0.78 | 0.68 | 0.89 | 0.69 | 0.62 | 0.77 | 0.66 | 0.6 | 0.72 |
| Analgesics/ analgesics | 1.71 | 1.4 | 2.08 | 1.21 | 1.02 | 1.45 | 0.96 | 0.82 | 1.12 | 0.8 | 0.7 | 0.91 | 0.83 | 0.74 | 0.92 |
| Analgesics/ drugs used in rheumatic diseases and gout | 1.59 | 1.08 | 2.36 | 0.97 | 0.59 | 1.58 | 0.58 | 0.43 | 0.78 | 0.5 | 0.38 | 0.67 | 0.53 | 0.41 | 0.69 |
| Anti-arrhythmic drugs | 0.73 | 0.46 | 1.18 | 0.69 | 0.47 | 1 | 0.62 | 0.46 | 0.85 | 0.43 | 0.31 | 0.58 | 0.44 | 0.34 | 0.58 |
| Anti-infective eye preparations | 1.79 | 1.34 | 2.4 | 1.25 | 0.9 | 1.74 | 0.85 | 0.63 | 1.16 | 0.79 | 0.63 | 1 | 0.72 | 0.6 | 0.87 |
| Anti-infective skin preparations | 1.75 | 1.28 | 2.39 | 0.99 | 0.66 | 1.5 | 0.9 | 0.66 | 1.22 | 0.69 | 0.54 | 0.89 | 0.64 | 0.52 | 0.78 |
| Antibacterial drugs | 2.07 | 1.82 | 2.36 | 1.26 | 1.1 | 1.44 | 0.92 | 0.81 | 1.04 | 0.87 | 0.79 | 0.96 | 0.87 | 0.81 | 0.95 |
| Antibacterial drugs/ acne and rosacea | 1.99 | 1.62 | 2.44 | 1.32 | 1.01 | 1.71 | 0.88 | 0.73 | 1.08 | 0.77 | 0.62 | 0.96 | 0.7 | 0.59 | 0.82 |
| Antibacterial drugs/ antiprotozoal drugs/ acne and rosacea | 2.56 | 1.96 | 3.34 | 1.17 | 0.92 | 1.49 | 1.08 | 0.8 | 1.44 | 0.61 | 0.47 | 0.81 | 0.79 | 0.64 | 0.97 |
| Anticoagulants and protamine | 0.69 | 0.54 | 0.88 | 0.44 | 0.35 | 0.55 | 0.53 | 0.43 | 0.66 | 0.45 | 0.37 | 0.54 | 0.47 | 0.41 | 0.55 |
| Antidepressant drugs | 2.44 | 2.01 | 2.96 | 1.07 | 0.84 | 1.35 | 0.98 | 0.8 | 1.21 | 0.78 | 0.66 | 0.93 | 0.73 | 0.63 | 0.85 |
| Antidepressant drugs/ analgesics/ analgesics | 1.27 | 0.92 | 1.76 | 1.13 | 0.86 | 1.48 | 0.7 | 0.55 | 0.89 | 0.5 | 0.41 | 0.6 | 0.63 | 0.53 | 0.76 |
| Antihistamine, hyposensitivity and allergic emergent | 1.92 | 1.51 | 2.43 | 1.1 | 0.83 | 1.46 | 0.95 | 0.76 | 1.2 | 0.63 | 0.53 | 0.75 | 0.6 | 0.51 | 0.71 |
| Antiplatelet drugs | 1.44 | 1.23 | 1.7 | 1.09 | 0.93 | 1.28 | 1.13 | 0.99 | 1.29 | 0.97 | 0.88 | 1.08 | 0.91 | 0.83 | 1 |

| Drug class | | | | | | | | | | | | | | |
|---|---|---|---|---|---|---|---|---|---|---|---|---|---|---|
| Antiprotozoal drugs/ drugs used in neuromuscular disorders | 0.79 | 0.46 | 1.35 | 0.79 | 0.51 | 1.21 | 0.63 | 0.44 | 0.91 | 0.5 | 0.38 | 0.64 | 0.59 | 0.47 | 0.73 |
| Antisecretory drugs and mucosal Protectants | 1.33 | 1.15 | 1.54 | 0.88 | 0.76 | 1.02 | 0.77 | 0.67 | 0.89 | 0.56 | 0.5 | 0.63 | 0.58 | 0.53 | 0.63 |
| Beta-adrenoceptor blocking drugs | 1.44 | 1.13 | 1.83 | 1.25 | 0.95 | 1.65 | 0.81 | 0.66 | 1 | 0.67 | 0.56 | 0.8 | 0.55 | 0.47 | 0.64 |
| Beta-adrenoceptor blocking drugs/ hypertension and heart failure/ stable angina, acute/ coronary syndrome and fibrin | 1.03 | 0.82 | 1.29 | 0.55 | 0.42 | 0.73 | 0.57 | 0.46 | 0.7 | 0.61 | 0.51 | 0.74 | 0.6 | 0.51 | 0.7 |
| Bronchodilators | 2.37 | 1.97 | 2.85 | 1.34 | 1.1 | 1.63 | 1.25 | 1.07 | 1.47 | 0.96 | 0.84 | 1.11 | 0.99 | 0.88 | 1.12 |
| Bronchodilators/ corticosteroids (respiratory) | 1.49 | 1.18 | 1.89 | 0.86 | 0.68 | 1.09 | 0.86 | 0.67 | 1.1 | 0.68 | 0.55 | 0.84 | 0.66 | 0.55 | 0.79 |
| Chronic bowel disorders/ corticosteroids (respiratory)/ antibacterial drugs/ antiprotozoal drugs/ corticosteroids (endocrine)/ drugs affecting the immune response/ drugs used in rheumatic diseases and gout | 2.94 | 2.41 | 3.6 | 1.5 | 1.21 | 1.84 | 1.51 | 1.25 | 1.81 | 1.13 | 0.97 | 1.3 | 1.08 | 0.95 | 1.23 |
| Corticosteroids and other anti-inflammatory preparations | 2.31 | 1.55 | 3.44 | 1.11 | 0.66 | 1.87 | 0.63 | 0.36 | 1.13 | 0.64 | 0.48 | 0.84 | 0.62 | 0.51 | 0.77 |
| Corticosteroids (respiratory) | 1.93 | 1.52 | 2.47 | 0.94 | 0.67 | 1.31 | 0.84 | 0.66 | 1.07 | 0.86 | 0.71 | 1.05 | 0.72 | 0.61 | 0.85 |
| Cough preparations | 2.25 | 1.65 | 3.08 | 1.36 | 0.92 | 2.02 | 1.27 | 0.96 | 1.68 | 0.84 | 0.64 | 1.09 | 0.91 | 0.73 | 1.13 |
| Diuretics | 1.17 | 1.01 | 1.36 | 0.93 | 0.77 | 1.12 | 0.91 | 0.8 | 1.03 | 0.78 | 0.7 | 0.87 | 0.76 | 0.69 | 0.83 |
| Diuretics/ minerals | 1.17 | 0.88 | 1.54 | 0.96 | 0.76 | 1.21 | 0.87 | 0.66 | 1.15 | 0.7 | 0.58 | 0.84 | 0.64 | 0.55 | 0.73 |
| Drugs acting on the ear | 2.1 | 1.47 | 3.01 | 1.78 | 1.29 | 2.45 | 1.05 | 0.76 | 1.46 | 0.79 | 0.63 | 0.99 | 0.83 | 0.66 | 1.04 |
| Drugs acting on the nose | 1.6 | 1.18 | 2.18 | 1.07 | 0.87 | 1.32 | 0.78 | 0.63 | 0.96 | 0.63 | 0.54 | 0.75 | 0.58 | 0.49 | 0.68 |
| Drugs acting on the oropharynx | 1.95 | 1.44 | 2.62 | 1.26 | 0.91 | 1.74 | 0.92 | 0.61 | 1.4 | 0.71 | 0.52 | 0.96 | 0.79 | 0.63 | 1 |
| Drugs affecting bone metabolism | 1.02 | 0.54 | 1.92 | 0.85 | 0.58 | 1.24 | 0.47 | 0.34 | 0.66 | 0.71 | 0.56 | 0.9 | 0.6 | 0.51 | 0.7 |
| Drugs for Genito-urinary disorders | 1.12 | 0.85 | 1.49 | 0.81 | 0.59 | 1.13 | 0.71 | 0.54 | 0.92 | 0.67 | 0.52 | 0.87 | 0.57 | 0.47 | 0.69 |
| Drugs used in diabetes | 0.82 | 0.68 | 0.99 | 0.87 | 0.7 | 1.08 | 0.65 | 0.53 | 0.79 | 0.58 | 0.5 | 0.67 | 0.56 | 0.49 | 0.65 |
| Drugs used in nausea and vertigo | 2.15 | 1.68 | 2.75 | 1.08 | 0.83 | 1.4 | 1.05 | 0.88 | 1.25 | 0.76 | 0.61 | 0.96 | 0.82 | 0.7 | 0.96 |

| Category | | | | | | | | | | | | | | | |
|---|---|---|---|---|---|---|---|---|---|---|---|---|---|---|---|
| Drugs used in psychoses and related disorders/ drugs used in nausea and vertigo | 1.52 | 1.15 | 1.99 | 0.97 | 0.69 | 1.36 | 0.83 | 0.61 | 1.12 | 0.69 | 0.47 | 1.02 | 0.53 | 0.4 | 0.69 |
| Drugs used in rheumatic diseases and gout | 2.06 | 1.71 | 2.48 | 1.16 | 0.97 | 1.4 | 0.76 | 0.64 | 0.91 | 0.76 | 0.65 | 0.88 | 0.73 | 0.64 | 0.82 |
| Dyspepsia and Gastro-oesophageal reflux disease | 2.43 | 1.73 | 3.4 | 1.13 | 0.91 | 1.41 | 0.87 | 0.69 | 1.11 | 0.62 | 0.5 | 0.78 | 0.7 | 0.6 | 0.83 |
| Emollient and barrier preparations | 1.64 | 1.24 | 2.15 | 1.16 | 0.92 | 1.47 | 0.87 | 0.73 | 1.05 | 0.8 | 0.69 | 0.93 | 0.76 | 0.68 | 0.84 |
| Hypertension and heart failure | 1.43 | 1.24 | 1.66 | 1.04 | 0.89 | 1.23 | 0.9 | 0.8 | 1.02 | 0.77 | 0.69 | 0.86 | 0.76 | 0.69 | 0.82 |
| Hypertension and heart failure/ drugs for Genito-urinary disorders | 0.97 | 0.72 | 1.32 | 0.92 | 0.71 | 1.2 | 0.79 | 0.63 | 0.99 | 0.58 | 0.47 | 0.72 | 0.56 | 0.48 | 0.65 |
| Hypnotics and anxiolytics | 2.26 | 1.67 | 3.06 | 0.86 | 0.58 | 1.27 | 1.27 | 0.99 | 1.63 | 0.95 | 0.73 | 1.25 | 0.8 | 0.63 | 1.01 |
| Hypnotics and anxiolytics/ antiepileptic drugs/ antiepileptic drugs/ drugs used in neuromuscular disorders/ general anaesthesia | 1.72 | 1.31 | 2.26 | 1 | 0.64 | 1.55 | 0.76 | 0.58 | 0.99 | 0.59 | 0.46 | 0.77 | 0.56 | 0.45 | 0.7 |
| Hypnotics and anxiolytics/ general anaesthesia | 1.77 | 1.25 | 2.52 | 0.89 | 0.51 | 1.54 | 0.94 | 0.64 | 1.39 | 0.89 | 0.7 | 1.13 | 0.72 | 0.56 | 0.92 |
| Laxatives | 1.58 | 1.22 | 2.05 | 1.42 | 1.09 | 1.86 | 0.96 | 0.8 | 1.15 | 0.9 | 0.79 | 1.03 | 0.75 | 0.67 | 0.85 |
| Lipid-regulating drugs | 1.36 | 1.17 | 1.58 | 0.98 | 0.83 | 1.16 | 0.94 | 0.83 | 1.08 | 0.73 | 0.66 | 0.81 | 0.72 | 0.65 | 0.8 |
| Local preparation for anal and rectal disorders | 1.56 | 1.14 | 2.15 | 1.03 | 0.66 | 1.6 | 0.81 | 0.59 | 1.1 | 0.65 | 0.51 | 0.83 | 0.47 | 0.38 | 0.59 |
| Miscellaneous Ophthalmic Preparations | 2.06 | 1.46 | 2.92 | 1.19 | 0.74 | 1.91 | 0.79 | 0.54 | 1.15 | 0.91 | 0.72 | 1.15 | 0.78 | 0.63 | 0.95 |
| Nitrates, calcium-channel blocker and other antianginal drugs | 1.39 | 1.2 | 1.62 | 1.16 | 0.98 | 1.36 | 1.09 | 0.95 | 1.24 | 0.87 | 0.78 | 0.97 | 0.87 | 0.79 | 0.95 |
| Oral nutrition | 1.68 | 0.73 | 3.83 | 1.05 | 0.63 | 1.73 | 0.72 | 0.46 | 1.15 | 1.09 | 0.77 | 1.53 | 1.15 | 0.88 | 1.5 |
| Positive inotropic drugs | 0.84 | 0.51 | 1.39 | 0.91 | 0.6 | 1.38 | 0.67 | 0.49 | 0.93 | 0.46 | 0.38 | 0.57 | 0.59 | 0.47 | 0.74 |
| Sex hormones | 1.32 | 0.95 | 1.84 | 0.76 | 0.49 | 1.19 | 0.72 | 0.52 | 1.01 | 0.64 | 0.43 | 0.96 | 0.55 | 0.42 | 0.71 |
| Soft-tissue disorders and topical pain relief | 2.37 | 1.72 | 3.26 | 1.88 | 1.45 | 2.43 | 1.19 | 0.91 | 1.55 | 0.87 | 0.71 | 1.06 | 0.76 | 0.66 | 0.88 |
| Thyroid and antithyroid drugs | 1.44 | 1.22 | 1.71 | 0.72 | 0.47 | 1.12 | 0.67 | 0.55 | 0.83 | 0.64 | 0.52 | 0.79 | 0.5 | 0.41 | 0.6 |
| Topical corticosteroids | 1.97 | 1.61 | 2.4 | 0.9 | 0.76 | 1.07 | 0.83 | 0.7 | 1 | 0.7 | 0.62 | 0.79 | 0.63 | 0.57 | 0.7 |
| Topical corticosteroids/ anti-infective skin preparations | 1.69 | 1.24 | 2.32 | 0.64 | 0.42 | 0.96 | 0.78 | 0.56 | 1.1 | 0.59 | 0.45 | 0.78 | 0.63 | 0.5 | 0.81 |
| Treatment of glaucoma | 1.04 | 0.57 | 1.87 | 0.97 | 0.5 | 1.89 | 0.73 | 0.44 | 1.2 | 0.8 | 0.56 | 1.15 | 0.61 | 0.47 | 0.79 |
| Vaccines and antisera | 1.85 | 1.57 | 2.19 | 1.23 | 1.01 | 1.49 | 0.91 | 0.78 | 1.07 | 0.75 | 0.66 | 0.85 | 0.74 | 0.67 | 0.81 |
| Vitamins | 0.96 | 0.74 | 1.24 | 0.89 | 0.64 | 1.25 | 0.68 | 0.51 | 0.9 | 0.52 | 0.43 | 0.64 | 0.51 | 0.42 | 0.61 |

S6 Table: Relative contribution scores for medically validated risk factors. In the columns, we have heart failure (HF) patient's average contribution scores, non-heart failure (Non-HF) patients average contribution scores, relevant standard deviation measures, and relative contribution (RC) and 95% confidence interval (CI) for these risk factors.

| Disease | HF | Standard Deviation: HF | Non-HF | Standard Deviation: Non-HF | RC | CI | |
|---|---|---|---|---|---|---|---|
| | | | | | | Lower bound | Upper bound |
| Myocardial infarction | 0.32 | 0.12 | 0.24 | 0.13 | 1.37 | 1.28 | 1.48 |
| Ischaemic stroke | 0.21 | 0.11 | 0.16 | 0.12 | 1.29 | 1.09 | 1.53 |
| Atrial fibrillation and flutter | 0.23 | 0.14 | 0.2 | 0.13 | 1.18 | 1.11 | 1.24 |
| Hypertension | 0.13 | 0.1 | 0.12 | 0.11 | 1.08 | 1.03 | 1.14 |
| Diabetes mellitus 1,2 | 0.12 | 0.09 | 0.12 | 0.11 | 1.02 | 0.9 | 1.15 |

S7 Table: Age-stratified, relative contribution (RC) and 95% confidence intervals (CI) for medically validated risk factors. In the columns, we have RC and CI for five age categories.

| Disease | 50-60 | | | | 60-65 | | | | 65-70 | | | | 70-75 | | | | 75-80 | | | |
|---|---|---|---|---|---|---|---|---|---|---|---|---|---|---|---|---|---|---|---|---|
| | RC | CI | | | RC | CI | | | RC | CI | | | RC | CI | | | RC | CI | | |
| | | LB | UB | | | LB | UB | | | LB | UB | | | LB | UB | | | LB | UB | |
| Atrial fibrillation and flutter | 1.23 | 1.02 | 1.47 | | 1.08 | 0.89 | 1.32 | | 1.3 | 1.15 | 1.47 | | 1.07 | 0.96 | 1.2 | | 1.02 | 0.93 | 1.12 | |
| Diabetes mellitus 1,2 | 1.48 | 1.05 | 2.07 | | 1.19 | 0.86 | 1.65 | | 0.84 | 0.64 | 1.1 | | 0.8 | 0.62 | 1.03 | | 0.7 | 0.56 | 0.89 | |
| Hypertension | 1.59 | 1.39 | 1.83 | | 1.21 | 1.05 | 1.4 | | 1.04 | 0.93 | 1.17 | | 0.83 | 0.75 | 0.92 | | 0.79 | 0.73 | 0.86 | |
| Ischaemic stroke | 1.86 | 1.05 | 3.3 | | 1.21 | 0.64 | 2.27 | | 0.84 | 0.57 | 1.24 | | 1.29 | 0.95 | 1.75 | | 1.17 | 0.88 | 1.54 | |
| Myocardial infarction | 1.55 | 1.33 | 1.8 | | 1.45 | 1.25 | 1.68 | | 1.36 | 1.13 | 1.63 | | 1.16 | 0.99 | 1.37 | | 1.29 | 1.09 | 1.53 | |

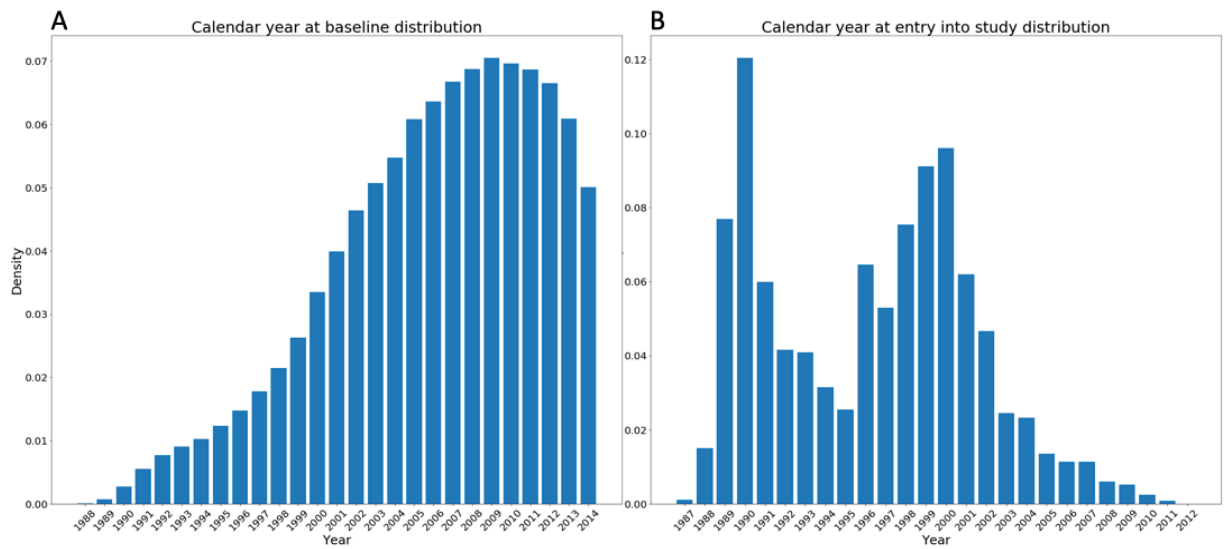

**Online Figure 1. Calendar year at baseline (A) and at entry into study (B) distributions.** A demonstrates the density of population at baseline calendar years ranging from 1988 to 2014. No patients exist at baseline prior to 1988. B shows density of our study population at various entry into study calendar years ranging from 1988 to 2012. The x axes for both A and B are calendar years while y, population density represented from [0,1].

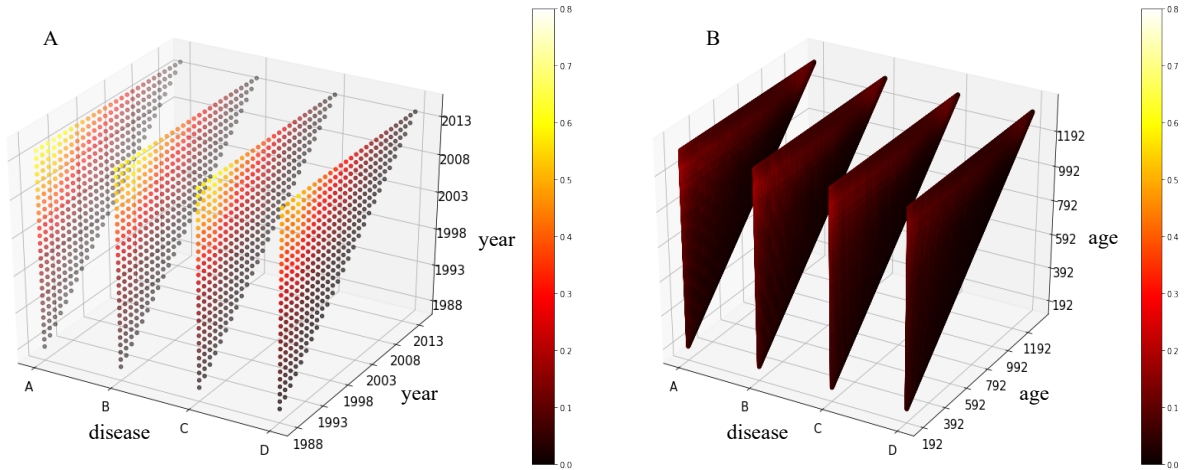

**Online Figure 2. Age and year embedding analysis.** We show cosine similarity measurement for summed embedding of diseases within different year (A) and age groups (B). A: depression, B: peripheral arterial disease, C: anxiety disorders, D: hypo or hyperthyroidism; age axis represents age in months from 16 to 100 years in months; year axis represents year from 1988 to 2014. Lighter colours indicate higher dissimilarity and darker colours lesser dissimilarity. The greater dissimilarity of summed embeddings by variation in year compared with variation in age suggests that year captures more information for prediction.

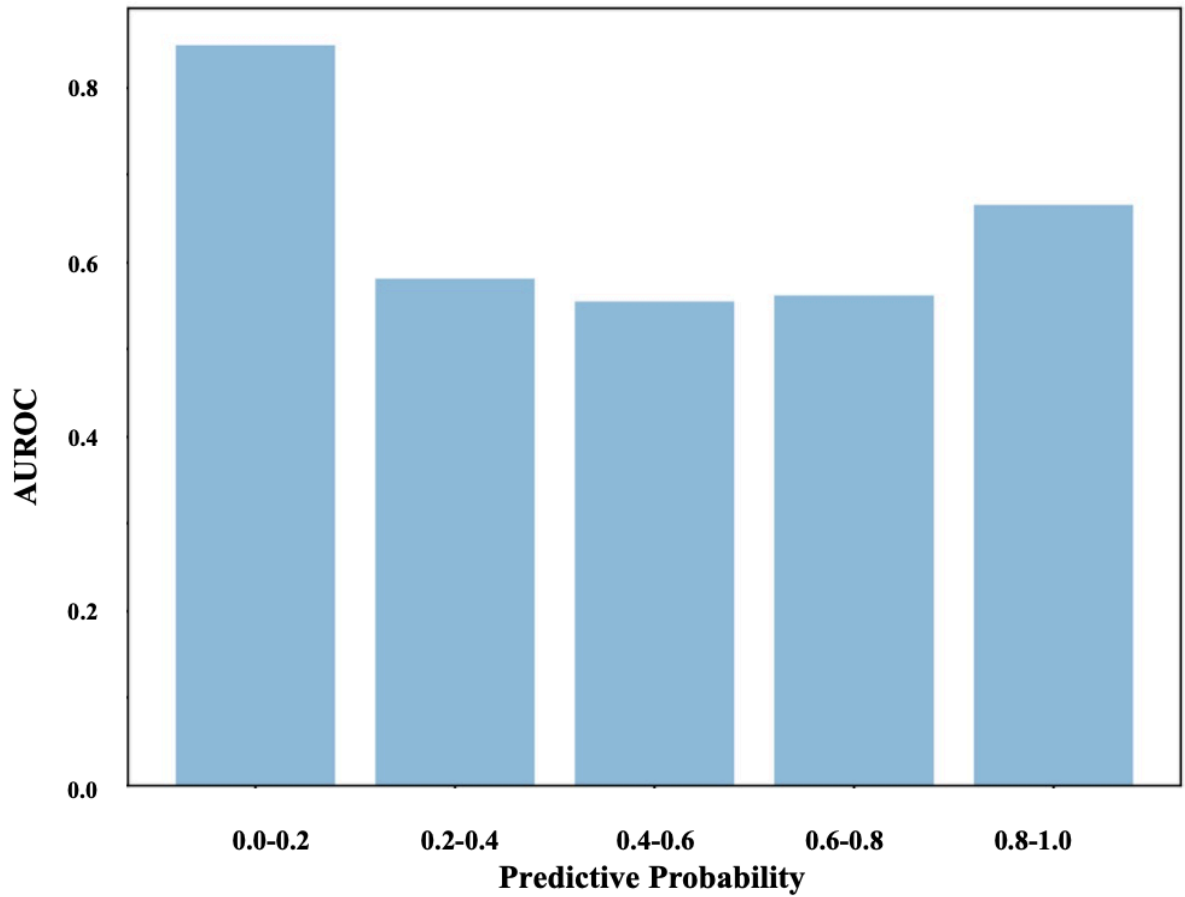

**Online Figure 3: Area Under the Receiver Operating Characteristic (AUROC) curve for the combined DMAY Validation and Test Dataset by quintiles of predicted probability**. The lowest performing strata is the [0.4-0.6) predictive probability strata and the best performing strata is the first one: [0,0.2).